\documentclass[10pt, a4paper]{article}

\usepackage[final]{lrec2026}
\usepackage{booktabs}
\usepackage[table]{xcolor}
\newcommand{\maxf}[1]{{\cellcolor[gray]{0.8}} #1}
\usepackage{bm}
\usepackage{bbding}
\usepackage{makecell}
\usepackage{siunitx}
\usepackage{multirow}
\usepackage[baseline]{euflag}
\newcommand{\tabitem}{~~\llap{\textbullet}~~}

\title{Same Meaning, Different Scores: Lexical and Syntactic Sensitivity in LLM Evaluation}

\name{Bogdan Kostić$^{*\dagger}$, Conor Fallon$^*$, Julian Risch$^\dagger$, Alexander Löser$^*$} 
\address{$^*$Berliner Hochschule für Technik (BHT) - Luxemburger Straße 10, 13467 Berlin \\
         $^\dagger$deepset GmbH - Zinnowitzer Straße 1, 10115 Berlin\\
         \{bogdan.kostic, julian.risch\}@deepset.ai, \{conor.fallon, aloeser\}@bht-berlin.de\\}

\abstract{
The rapid advancement of Large Language Models (LLMs) has established standardized evaluation benchmarks as the primary instrument for model comparison. Yet, their reliability is increasingly questioned due to sensitivity to shallow variations in input prompts.
This paper examines how controlled, truth-conditionally equivalent lexical and syntactic perturbations affect the absolute performance and relative ranking of 23 contemporary LLMs across three benchmarks: MMLU, SQuAD, and AMEGA.
We employ two linguistically principled pipelines to generate meaning-preserving variations: one performing synonym substitution for lexical changes, and another using dependency parsing to determine applicable syntactic transformations.
Results show that lexical perturbations consistently induce substantial, statistically significant performance degradation across nearly all models and tasks, while syntactic perturbations have more heterogeneous effects, occasionally improving results.
Both perturbation types destabilize model leaderboards on complex tasks.
Furthermore, model robustness did not consistently scale with model size, revealing strong task dependence.
Overall, the findings suggest that LLMs rely more on surface-level lexical patterns than on abstract linguistic competence, underscoring the need for robustness testing as a standard component of LLM evaluation.
 \\ \newline  }

\begin{document}

\maketitleabstract

\section{Introduction}
\begin{figure*}
    \centering
    \includegraphics[width=\linewidth]{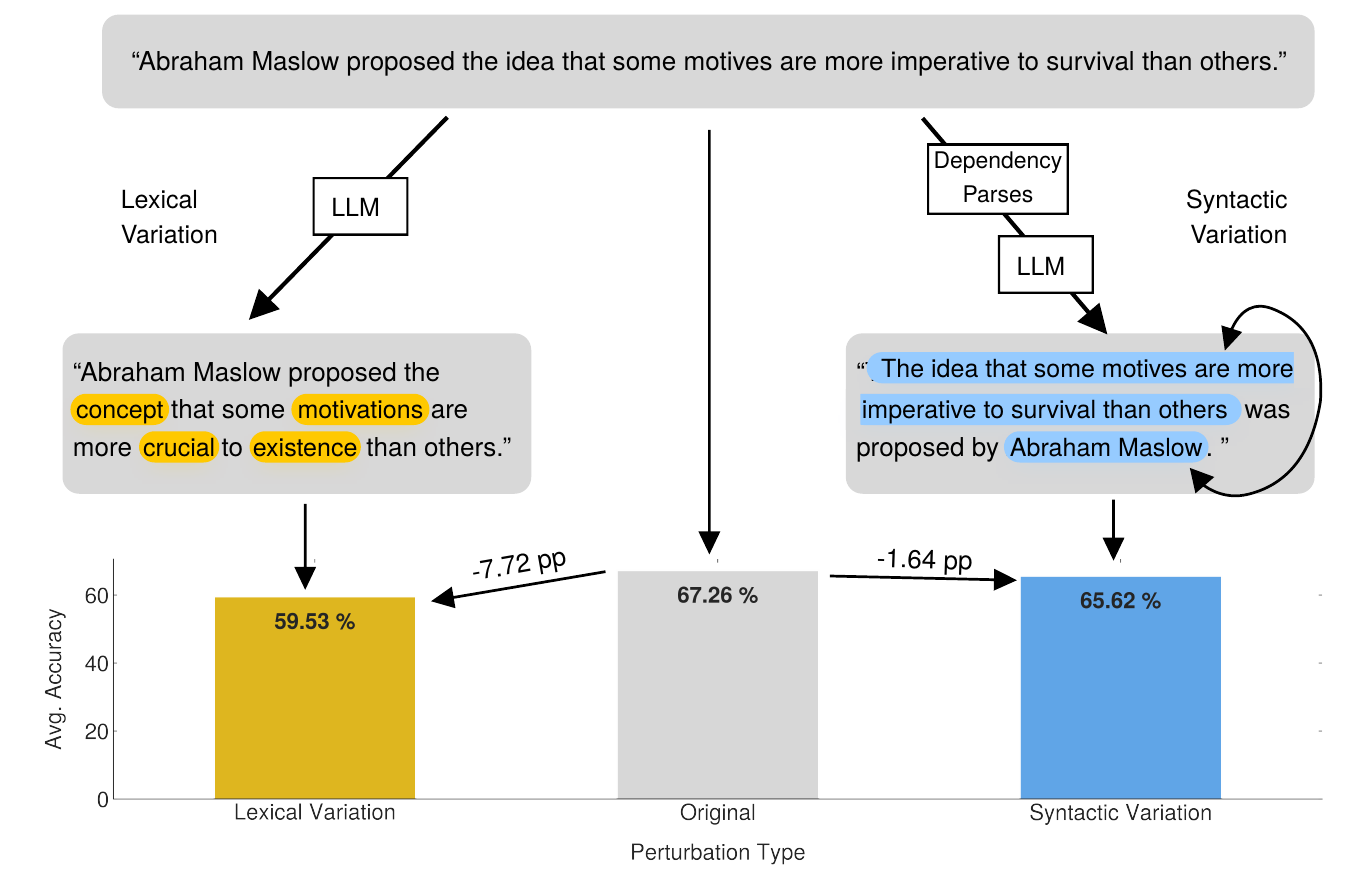}
    \caption{Illustration of the two linguistically principled, meaning-preserving perturbation pipelines and their average impact on LLM performance on the MMLU benchmark. Yellow and blue annotations mark changed words and moved constituents, respectively. The bar chart quantifies one of the core findings: lexical perturbations induce a substantial average accuracy drop, while the impact of syntactic perturbations is smaller.}
    \label{fig:fig_1}
\end{figure*}
The rapid proliferation of new LLMs has established standardized evaluation benchmarks as the primary tool for comparison, making public leaderboards the default instrument for model selection.
There is a growing concern that such metrics may overestimate the true generalization capabilities of LLMs, partly due to data leakage from benchmark test sets into vast, undisclosed training corpora, and more fundamentally, due to an observable sensitivity to superficial variations in the input prompt.
Prior work demonstrated that model performance can degrade significantly with the introduction of irrelevant context, the reordering of premises in reasoning tasks, or even minor alterations to the symbols used for answer choices in multiple-choice questions.
Yet, these studies often do not distinguish between different kinds of linguistic changes, leaving a key question unanswered: Is the observed performance variance driven more by lexical shifts or by alterations in syntactic structure?

This paper concentrates specifically on the impact of meaning-preserving lexical and syntactic variations on LLM performance.
We investigate how controlled, truth-conditionally equivalent perturbations at the lexical and syntactic levels affect the absolute performance and relative ranking of modern LLMs across diverse tasks.
This provides a robust empirical basis for assessing the stability of model evaluation and the reliability of benchmark-driven leaderboards.
Our investigation is motivated by the need to move beyond static benchmark scores toward more reliable assessments of the true generalization capabilities of LLMs.
Such an approach aligns with the principles of veridical data science, which posits stability, the robustness of a model's results to reasonable variations, as a core pillar for establishing the trustworthiness of data-based models~\citep{yu_veridical_2024,alaa_veridical_2024}.

Concretely, we first select a \textbf{diverse set of three benchmark datasets}: MMLU for multiple-choice question answering (QA), SQuAD for extractive QA, and AMEGA for clinical guideline adherence.
We then systematically perturb each dataset using two distinct, linguistically-grounded perturbation pipelines, as exemplified in Figure~\ref{fig:fig_1}.
The lexical pipeline uses an LLM to perform guided synonym substitution, ensuring contextual appropriateness, while the syntactic pipeline leverages dependency parsing to identify applicable grammatical transformations, which are subsequently executed by an LLM.
Second, we evaluate a cohort of \textbf{23 state-of-the-art LLMs} varying in size, architecture, provider, and accessibility on both the original and perturbed versions of each benchmark.
Third, we \textbf{quantify the resulting performance variance} using task-appropriate metrics and apply statistical tests to assess the stability of absolute scores and relative rankings.

Our three main contributions are: 
\textbf{(1) LLMs rely on surface lexical cues over abstract syntactic structure.} We find that models are significantly more sensitive to lexical variation than to syntactic rephrasing.
\textbf{(2) Benchmark leaderboards are brittle.} We demonstrate that minor, meaning-preserving changes are sufficient to alter model rankings significantly, challenging the reliability of static leaderboards as a tool for model selection.
\textbf{(3) Model scaling does not confer enhanced robustness.} We find that bigger models are not necessarily more robust: the relationship between model size and stability is not monotonic and task-dependent.
We release our implementation and the perturbed benchmark datasets on GitHub to support reproducibility and follow-up work.\footnote{\url{https://github.com/bogdankostic/llm-prompt-perturbation}}

\section{Related Work}

Prior work examines the stability of LLMs by probing how small, meaning-preserving changes to prompts alter model behavior, typically through controlled perturbation families on specific downstream tasks.
Across settings, results suggest a common pattern: seemingly innocuous differences in context, order, formatting, or wording can induce large shifts in performance, model rankings, and generated content.

\textbf{Irrelevant context degrades performance.} Adding extraneous information to a prompt can substantially degrade LLM performance.
For instance, injecting unrelated text into grade-school math problems causes significant accuracy drops, suggesting LLMs struggle to filter distractions and isolate task-relevant content~\citep{shi_large_2023}.
Moreover, perturbing medical question-answering benchmarks by adding irrelevant and biased patient characteristics reveals that some models are highly sensitive to stereotyped information that would not mislead human clinicians~\citep{ness_medfuzz_2024}.

\textbf{Order \& Formatting vs. Performance.}
LLM performance is also sensitive to the order and format of information.
Studies show this sensitivity applies to the permutation of logical premises \citep{chen_premise_2024}, the position of key documents in the context in open-book QA settings \citep{liu_lost_2024}, and the ordering of few-shot examples in the prompt \citep{lu_fantastically_2022}.
In multiple-choice settings, performance and model rankings are affected by the order of answer choices and choice of option labels \citep{brucks_prompt_2025, pezeshkpour_large_2024, alzahrani_when_2024, zheng_large_2023}.
In addition, pure formatting perturbations, such as casing, whitespace, and enumeration wrappers, lead to a significant performance spread on various multiple-choice and classification tasks, regardless of model size~\citep{sclar_quantifying_2023}.

\textbf{Intra-Sentence Level Granularity.}
Building on these findings, our work examines order sensitivity at a finer granularity: within individual sentences.
Prior studies mostly consider rearrangements of sentences or larger context chunks.
In contrast, we hold truth-conditional content fixed while permuting the order of constituents inside sentences.
This design allows us to assess whether valid, meaning-preserving intra-sentential permutations influence LLM performance, complementing prior results on permutations of larger units.

\textbf{Word-level perturbations \& paraphrasing.}
According to numerous studies, meaning-preserving word-level edits and paraphrases can substantially impact LLM behavior as well.
\citet{mizrahi_state_2024} demonstrate through large-scale multi-prompt evaluations that simply paraphrasing task instructions within a prompt can flip relative rankings across models.
Similarly, \citet{zhu_promptrobust_2024} investigate how different character-level, word-level, sentence-level, and semantic-level perturbations of the instructions can impact performance across a range of tasks and conclude that word-level perturbations have the most significant effect on task performance.
\citet{mirzadeh_gsm-symbolic_2024} investigate the generalization capabilities for mathematical reasoning tasks by comparing template-based variants of grade school mathematical problems and find strong sensitivity to surface changes, especially when numeric entities are modified.
For intent classification and slot filling, \citet{qiang_prompt_2024} report synonym substitutions, oronyms, and general paraphrasing to reduce accuracy.
Complementing these findings, \citet{ackerman_novel_2024} show that paraphrasing typically causes larger performance shifts than superficial modifications like punctuation or casing.
Similarly, controlled logic tasks reveal significant drops from minor lexical tweaks, suggesting that models may lean on token-specific cues rather than stable reasoning patterns~\citep{jiang_peek_2024}.
\citet{zhao-etal-2024-improving} propose a two-stage training framework, including consistency alignment, specifically to improve model robustness against such variations.

Extending robustness beyond synonymy and paraphrasing, \citet{zheng_neo-bench_2024} evaluate the response of LLMs to language change and identify a severe performance degradation when prompts contain neologisms.
\citet{gan_reasoning_2024} find that introducing minor typographical errors into benchmark questions yields significant accuracy declines.
A plausible explanation is the ``\emph{curse of tokenization}'': subword segmentation leads to the mapping of near-identical strings to different token sequences, such that minor edits change token boundaries and are propagated throughout the computation~\citep{chai_tokenization_2024}.

\textbf{Our work} extends prior robustness studies that show LLM performance shifts under irrelevant context, reordering, formatting, and paraphrasing by introducing linguistically principled, truth-conditionally equivalent perturbations at both lexical and syntactic levels, generated using controlled synonym substitution and dependency-guided transformations.
We further present a novel analysis of the task-dependent relationship between model size and robustness.

\section{Methodology}
\begin{figure*}
    \centering
    \includegraphics[width=\linewidth]{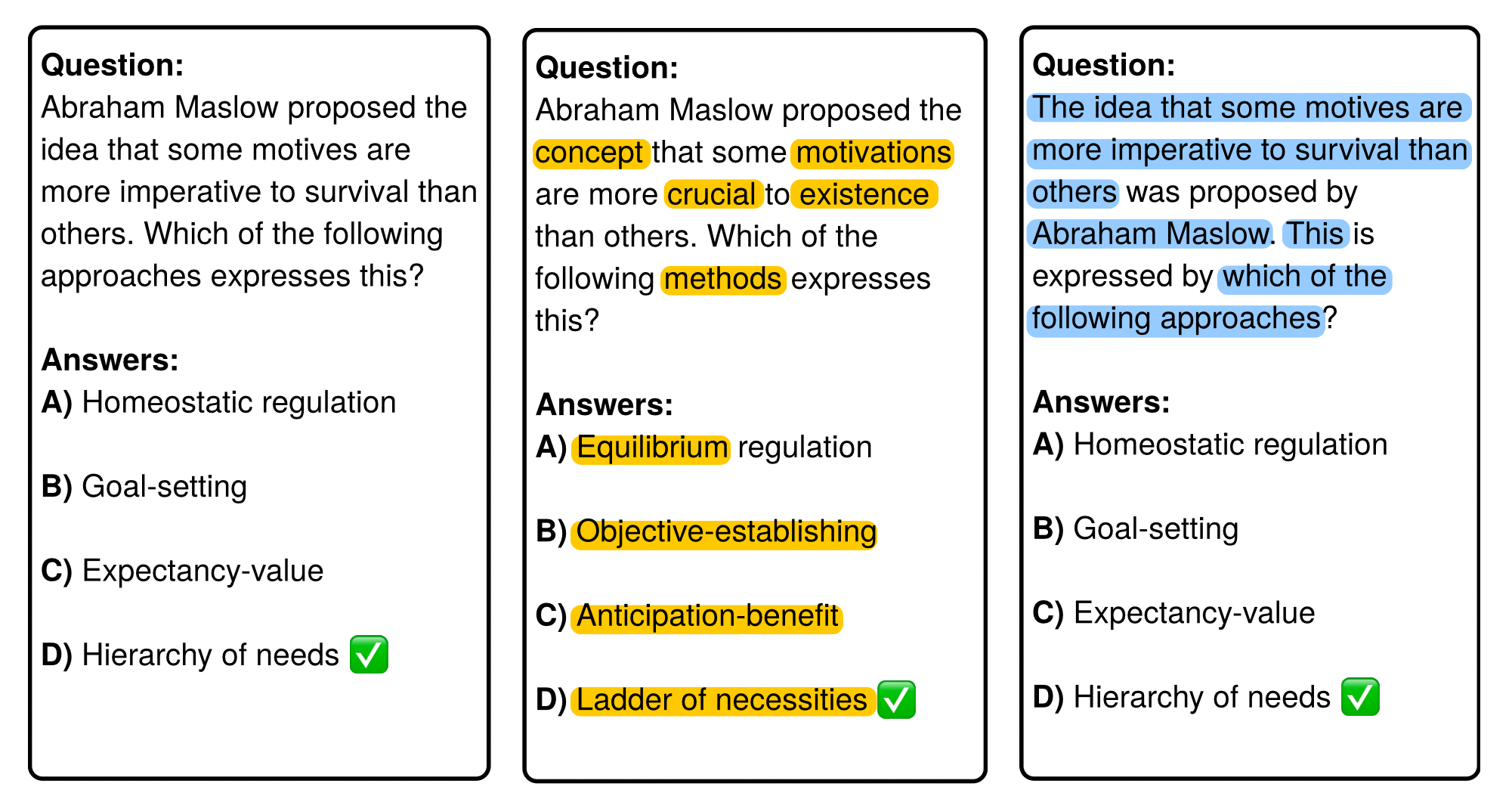}
    \caption{Example from MMLU: original item (left), lexically perturbed version (center), and syntactically perturbed version (right). Changed words are marked in yellow and moved constituents in blue.}
    \label{fig:example}
\end{figure*}
Our methodology involves three stages: selecting diverse benchmarks and models, applying linguistically-principled perturbations, and conducting a rigorous evaluation.
We first describe the chosen benchmarks and models, then detail the lexical and syntactic perturbation generation methods, and finally outline the evaluation protocol.

\begin{table}
\setlength{\tabcolsep}{5pt}
\centering
\small
\begin{tabular}{llll}
\toprule
\textbf{} & \textbf{MMLU} & \textbf{SQuAD} & \textbf{AMEGA} \\
\midrule
\textbf{Task}
& \makecell[l]{Multiple-choice\\QA}
& \makecell[l]{Extractive\\QA}
& \makecell[l]{Free-form\\QA} \\
\addlinespace[3pt]
\textbf{Domain}
& \makecell[l]{Humanities,\\social sciences,\\STEM + others}
& \makecell[l]{Wikipedia}
& \makecell[l]{Clinical} \\
\addlinespace[3pt]
\textbf{Size}
& \makecell[l]{15,858\\questions}
& \makecell[l]{1,000\\questions}
& \makecell[l]{135 ques-\\tions with\\1,337 eval.\\criteria} \\
\addlinespace[3pt]
\textbf{Metric}
& \makecell[l]{Accuracy}
& \makecell[l]{EM, F1 \&\\SAS}
& \makecell[l]{Guideline\\adherence\\score} \\

\bottomrule
\end{tabular}
\caption[Selected benchmarks for assessing LLM robustness to linguistic perturbations.]{Diverse representative benchmarks to assess LLM robustness to linguistic perturbations.}
\label{tab:benchmarks}
\end{table}

\subsection{Benchmark Datasets}
We select three diverse benchmarks to cover a range of tasks and domains (summarized in Table~\ref{tab:benchmarks}).
\textbf{MMLU} \citep{hendrycks_measuring_2020} is a 14,000-question test covering knowledge from humanities, STEM, social sciences, and more, in a multiple-choice QA format.
It assesses broad world knowledge and reasoning.
\textbf{SQuAD} \citep{rajpurkar_squad_2016} is a standard extractive QA dataset with questions on Wikipedia articles.
Models must locate the answer span in the provided passages.
We use a representative sampled subset of 1,000 questions.
\textbf{AMEGA} \citep{fast_autonomous_2024} is a clinical domain benchmark requiring free-form answers to medical questions, with an automatic evaluation of guideline adherence.
It comprises 20 physician-crafted diagnostic cases spanning 13 specialties, featuring 135 open-ended questions on these cases and 1,337 weighted scoring criteria.
Each model-generated answer is scored against a set of criteria by an LLM-based evaluator, with a maximum of 50 points per clinical case.
These three benchmarks enable testing robustness in contexts of factual knowledge recall (MMLU), reading comprehension (SQuAD), and specialized reasoning under guidelines (AMEGA).
Appropriate task metrics are used for evaluation: accuracy for MMLU, exact match, F1, and semantic answer similarity for SQuAD, and guideline adherence score for AMEGA.

\subsection{Evaluated Models}
We evaluate a cohort of 23 state-of-the-art LLMs that vary in size (from $\sim$0.3B up to $\sim$235B parameters), architecture (standard dense transformers vs.~mixture-of-experts), and availability (open-weight vs.~proprietary).
All models are accessed through a unified interface provided by the open-source framework Haystack \citep{pietsch_haystack_2019} to ensure consistent prompt formatting.
Each model is evaluated on every original and perturbed dataset instance.
This broad coverage allows us to analyze whether certain model attributes, such as size, correlate with robustness.
Importantly, all models are tested in a zero-shot setting to assess their capabilities and sensitivity to prompt phrasing directly.
All evaluations are conducted using NVIDIA A100 GPUs for the open-weight models. 
To ensure deterministic and comparable outputs, we set the model temperature to 0 and used a fixed random seed for both the perturbation and generation processes.

\begin{table*}[t]
\scriptsize
\centering
\setlength{\tabcolsep}{3.0pt}
\begin{tabular}{l
S S[table-format=2.1,table-space-text-post={$^{***}$}] S[table-format=1.1,table-space-text-post={$^{***}$}]
S[table-format=2.1,table-space-text-post=~/~$83.7$~/~$91.0$] S[table-format=1.1,table-space-text-post=$^{***}$/~$3.7$~/~$1.0$] S[table-format=1.1,table-space-text-post=$^{***}$/~$3.7$~/~$1.0$] 
S S[table-format=1.1,table-space-text-post={$^{***}$}] S[table-format=1.1,table-space-text-post={$^{***}$}]}
\toprule
& \multicolumn{3}{c}{\textbf{MMLU}} & \multicolumn{3}{c}{\textbf{SQuAD}~(EM~/~F1~/~SAS)} & \multicolumn{3}{c}{\textbf{AMEGA}}\\
\cmidrule(lr){2-4}\cmidrule(lr){5-7}\cmidrule(lr){8-10}
\multicolumn{1}{c}{Model} & {Orig} & {$\Delta$ Lex} & {$\Delta$ Syn} & \multicolumn{1}{c}{Orig} & \multicolumn{1}{c}{$\Delta$ Lex} & \multicolumn{1}{c}{$\Delta$ Syn} & {Orig} & {$\Delta$ Lex} & {$\Delta$ Syn} \\
\midrule
GPT-5-Nano & 69.6 & \maxf{10.2$^{***}$} & 2.2$^{***}$ & 66.4~/~$83.7$~/~$91.0$ & 3.5$^{***}$/~$3.6$~/~$2.2$ & 3.2$^{*\hphantom{**}}$/~$2.6$~/~$1.7$ & 37.4 & 1.9$^{**}$ & 0.6 \\\addlinespace[1pt]
GPT-5-mini & 80.0 & 9.3$^{***}$ & 1.8$^{***}$ & 74.9~/~$89.5$~/~$93.7$ & 4.0$^{***}$/~$3.5$~/~$1.6$ & 3.0$^{*\hphantom{**}}$/~$2.4$~/~$1.3$ & 39.6 & 2.1$^{***}$ & 1.6$^{**}$ \\\addlinespace[1pt]
GPT-4.1-Nano & 69.9 & 7.8$^{***}$ & 2.0$^{***}$ & 76.8~/~$89.7$~/~$94.1$ & 5.3$^{***}$/~$4.0$~/~$2.1$ & 3.5$^{***}$/~$3.0$~/~$1.5$ & 34.1 & 0.7 & 0.6 \\\addlinespace[1pt]
GPT-4.1-mini & 80.7 & 8.3$^{***}$ & 1.6$^{***}$ & 77.2~/~$90.7$~/~$94.4$ & 4.5$^{***}$/~$3.7$~/~$1.9$ & 2.0$^{*\hphantom{**}}$/~$2.3$~/~$1.3$ & 36.0 & 0.3 & 0.9 \\\addlinespace[1pt]
GPT-OSS-120b &  \maxf{86.1} & 9.7$^{***}$ & 2.4$^{***}$ & 71.9~/~$87.3$~/~$93.5$ & 4.4$^{***}$/~$3.4$~/~$1.9$ & 1.6$^{\hphantom{***}}$/~$2.0$~/~$0.9$ & \maxf{39.8} & 0.5 & 0.7 \\\addlinespace[1pt]
GPT-OSS-20b & 81.4 & 9.6$^{***}$ & 2.3$^{***}$ & 70.8~/~$87.3$~/~$92.5$ & 4.8$^{***}$/~$4.0$~/~$1.8$ & 1.9$^{\hphantom{***}}$/~$2.0$~/~$0.3$ & 37.7 & 1.7$^{*}$ & 0.0 \\\addlinespace[1pt]
Llama-3.3-70B-Instruct & 80.4 & 9.3$^{***}$ & 1.6$^{***}$ & 82.3~/~$92.6$~/~$95.9$ & 5.1$^{***}$/~$3.5$~/~$1.9$ & 3.0$^{***}$/~$2.3$~/~$1.4$ & 32.7 & 0.5 & 0.4 \\\addlinespace[1pt]
Llama-3.1-8B-Instruct & 62.8 & 8.1$^{***}$ & 2.3$^{***}$ & 72.3~/~$86.6$~/~$92.0$ & 5.3$^{***}$/~$3.4$~/~$1.6$ & 2.6$^{*\hphantom{**}}$/~$2.1$~/~$1.3$ & 29.8 & 1.5$^{*}$ & 0.0 \\\addlinespace[1pt]
Llama-3.2-3B-Instruct & 58.0 & 7.0$^{***}$ & 2.0$^{***}$ & 75.4~/~$87.2$~/~$92.8$ & 3.4$^{**\hphantom{*}}$/~$3.1$~/~$1.5$ & 1.8$^{\hphantom{***}}$/~$2.6$~/~$1.7$ & 26.6 & 1.3 & -0.8 \\\addlinespace[1pt]
Llama-3.2-1B-Instruct & 25.7 & 0.9$^{***}$ & 0.1 & 57.8~/~$71.5$~/~$84.4$ & 4.2$^{***}$/~$4.1$~/~$2.5$ & 2.6$^{**\hphantom{*}}$/~$3.9$~/~$2.6$ & 22.2 & 2.5$^{**\hphantom{*}}$ & 0.4 \\\addlinespace[1pt]
gemini-2.5-flash & 84.9 & 8.8$^{***}$ & 2.1$^{***}$ & 86.6~/~$94.3$~/~$96.8$ & 2.9$^{**\hphantom{*}}$/~$2.5$~/~$1.2$ & 2.9$^{***}$/~$2.3$~/~$1.2$ & 38.1 & 0.1 & 1.1 \\\addlinespace[1pt]
gemini-2.5-flash-lite & 63.5 & 8.6$^{***}$ & 0.3 & 85.6~/~$93.4$~/~$96.1$ & 5.3$^{***}$/~$3.9$~/~$1.8$ & 3.1$^{***}$/~$2.8$~/~$1.7$ & 36.0 & 1.1 & 1.6$^{**}$ \\\addlinespace[1pt]
gemma-3-27b-it & 76.5 & 9.5$^{***}$ & 1.6$^{***}$ & 76.1~/~$90.1$~/~$94.0$ & 4.3$^{***}$/~$2.9$~/~$1.5$ & 3.0$^{***}$/~$2.8$~/~$1.7$ & 35.4 & 1.2 & -0.1 \\\addlinespace[1pt]
gemma-3-12b-it & 71.2 & 8.7$^{***}$ & 2.4$^{***}$ & 81.9~/~$91.5$~/~$95.1$ & 4.2$^{***}$/~$2.9$~/~$1.5$ & 4.2$^{***}$/~$3.0$~/~$1.8$ & 35.1 & 0.7 & -0.7 \\\addlinespace[1pt]
gemma-3-4b-it & 57.2 & 5.7$^{***}$ & 1.3$^{**}$ & 78.8~/~$89.6$~/~$93.7$ & 4.1$^{***}$/~$3.0$~/~$1.5$ & 3.6$^{***}$/~$3.3$~/~$2.1$ & 32.7 & 1.2 & 0.6 \\\addlinespace[1pt]
gemma-3-1b-it & 38.8 & 2.7$^{***}$ & 0.8 & 64.7~/~$77.6$~/~$87.3$ & 4.2$^{***}$/~$4.4$~/~$2.1$ & \maxf{5.4$^{***}$/~$5.3$~/~$3.1$} & 26.9 & 0.7 & 0.9 \\\addlinespace[1pt]
gemma-3-270m-it & 13.3 & -0.3 & -0.5 & 20.8~/~$36.0$~/~$61.2$ & 0.9$^{\hphantom{***}}$/~$2.6$~/~$2.0$ & 3.8$^{***}$/~$5.8$~/~$4.5$ & 15.7 & \maxf{4.0$^{***}$} & 1.3$^{*}$ \\\addlinespace[1pt]
Mistral-Large-Instruct-2411 & 80.0 & 9.4$^{***}$ & 2.0$^{***}$ & \maxf{87.0~/~$93.4$~/~$96.6$} & 3.8$^{***}$/~$2.8$~/~$1.0$ & 1.8$^{*\hphantom{**}}$/~$1.7$~/~$1.2$ & 34.9 & 2.4$^{***}$ & 1.1$^{*}$ \\\addlinespace[1pt]
Mistral-Small-3.2-24B-Instruct-2506 & 76.6 & 9.3$^{***}$ & 2.1$^{***}$ & 84.2~/~$92.9$~/~$95.8$ & 5.1$^{***}$/~$3.4$~/~$1.9$ & 1.9$^{*\hphantom{**}}$/~$1.9$~/~$1.2$ & 36.8 & 1.5$^{*}$ & \maxf{1.7$^{**}$} \\\addlinespace[1pt]
Ministral-8B-Instruct-2410 & 61.7 & 7.6$^{***}$ & \maxf{2.4$^{***}$} & 79.9~/~$89.0$~/~$93.4$ & 4.2$^{***}$/~$3.7$~/~$1.8$ & 0.6$^{\hphantom{***}}$/~$1.9$~/~$1.1$ & 30.8 & 1.5$^{*}$ & 0.9 \\\addlinespace[1pt]
Qwen3-235B-A22B-Instruct-2507 & 85.5 & 9.4$^{***}$ & 1.9$^{***}$ & 71.1~/~$85.9$~/~$92.2$ & 2.1$^{\hphantom{***}}$/~$1.9$~/~$1.2$ & 1.5$^{\hphantom{***}}$/~$1.8$~/~$1.4$ & 37.3 & 0.3 & 0.1 \\\addlinespace[1pt]
Qwen3-30B-A3B-Instruct-2507 & 77.2 & 8.4$^{***}$ & 1.4$^{***}$ & 76.2~/~$88.9$~/~$93.8$ & 4.4$^{***}$/~$3.0$~/~$1.7$ & 2.3$^{*\hphantom{**}}$/~$2.3$~/~$1.3$ & 36.7 & 0.1 & -0.2 \\\addlinespace[1pt]
Qwen3-4B-Instruct-2507 & 66.0 & 9.7$^{***}$ & 1.7$^{***}$ & 69.3~/~$85.2$~/~$91.5$ & \maxf{5.8$^{***}$/~$4.4$~/~$2.6$} & 1.7$^{\hphantom{***}}$/~$2.7$~/~$2.3$ & 34.2 & 1.0 & -0.1 \\\addlinespace[1pt]
\bottomrule
\end{tabular}
\caption{LLM performance variance under lexical and syntactic perturbations across MMLU, SQuAD, and AMEGA. For each model, we report the original score and the absolute change under lexical and syntactic variants ($\Delta$ Lex, $\Delta$ Syn). MMLU uses accuracy, SQuAD uses EM~/~F1~/~SAS, and AMEGA uses the guideline adherence score out of 50 as metric. Positive $\Delta$ values indicate a drop in performance, and negative values indicate improvements. Grey backgrounds mark the highest value in each column. Asterisks mark significance (*: $p<.05$, **: $p<.01$, ***: $p<.001$) according to McNemar's test.}
\label{tab:results}
\end{table*}
\subsection{Meaning-Preserving Perturbations}

We generated perturbations that are lexically or syntactically different from the original but preserve its truth-conditional meaning.

\paragraph{Lexical Variation.}
To obtain natural-sounding lexical variation, we rewrote entire data instances through guided synonym substitution by a quantized version of \texttt{Llama-3.3-70B-Instruct}.
The prompt instructed the model to replace words with semantically appropriate synonyms while preserving meaning and domain-specific terminology.
For SQuAD, a crucial constraint was added to leave the contiguous answer string in the text passage unaltered.
Beyond free-form prompting, we constrained decoding with a JSON schema that returned the perturbed text and a \texttt{changes} list of \texttt{(original, substitution)} tuples, and we used benchmark-specific templates to enforce meaning preservation and consistent formatting across datasets.

\paragraph{Syntactic Variation.}
To alter sentence structure without modifying words, we applied a two-stage LLM-powered pipeline.
First, we segment inputs into sentences and parse them with spaCy's \texttt{en\_core\_web\_trf} model \cite{montani_spacy_2023} to identify matrix clause constituents such as subjects, objects, clausal complements, and expletives.
Transformations are only attempted when rule-based applicability conditions are met, sampling uniformly when multiple apply.
Perturbations target classic syntactic alternations, such as active-to-passive, extraposition, and wh-movement.
A complete list of transformations and the corresponding applicability conditions is provided in Appendix~\ref{sec:appendix_syntactic_transformations}.

The second stage prompts a quantized \texttt{Llama-3.3-70B-Instruct} with the original sentence, the selected transformation, and the extracted constituents, using operation-specific templates that enforce grammaticality and preservation of propositional content.
Edits are restricted to the matrix clause, i.e., embedded clauses remain unchanged.

Examples of both perturbation types on the MMLU dataset are presented in Figure~\ref{fig:example}.

\subsection{Evaluation Metrics}
Each model is evaluated on the original and perturbed versions of each dataset.
For MMLU, we use exact-match accuracy since each question has a single correct choice.
For SQuAD, we report the standard exact match and token-level F1 scores, as well as Semantic Answer Similarity (SAS) \citep{risch_semantic_2021}, which gives partial credit for paraphrased answers.
For AMEGA, model answers are scored by an automatic evaluator against the guideline criteria, yielding an overall adherence score out of 50.
We compute the score differences for each model between original and perturbed data, and use statistical tests to assess significance.
For each model individually, we use McNemar's test \citep{mcnemar_note_1947} to determine whether the proportion of correct predictions differs significantly between original and perturbed data.
To identify whether the distribution across all models between the scores on the original and the perturbed versions differs significantly, we apply the Wilcoxon signed-rank test \citep{wilcoxon_individual_1945}.
We also examine the stability of model rankings by computing Kendall’s $\tau$ rank correlation between the original and perturbed leaderboard ordering \citep{kendall_new_1938}.
We define two rankings to be strictly equivalent if $\tau > 0.9$ and moderately equivalent if $\tau > 0.8$.

\section{Results}

\begin{figure}
    \centering
    \includegraphics[width=\linewidth]{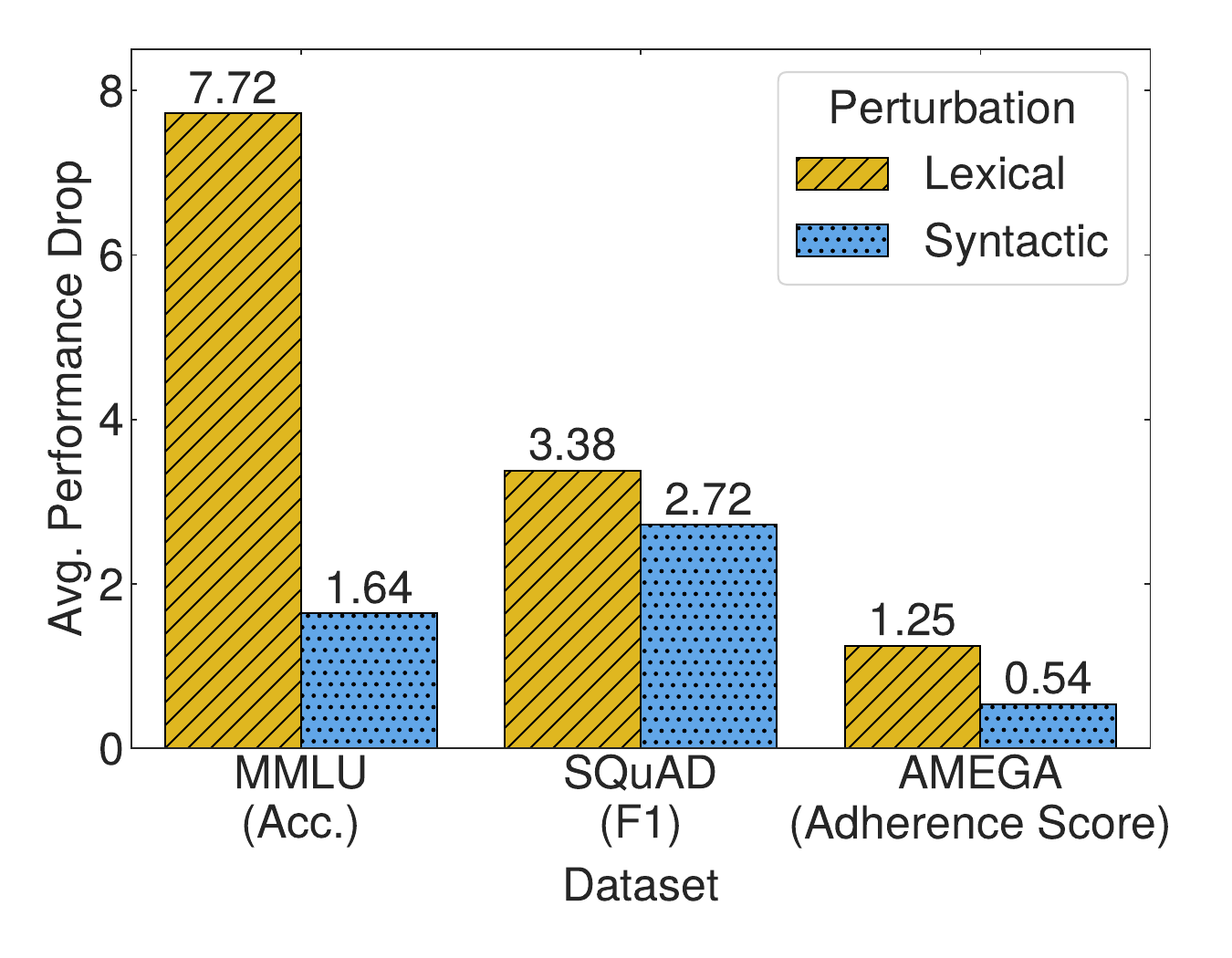}
    \vspace{-0.8cm}
    \caption{Average drop in performance after lexical and syntactic perturbation across 23 LLMs for MMLU, SQuAD, and AMEGA. Lexical perturbations cause larger drops, most notably on MMLU.}
    \label{fig:avg_performance_drop}
\end{figure}

\begin{figure*}
    \centering
    \includegraphics[width=\linewidth]{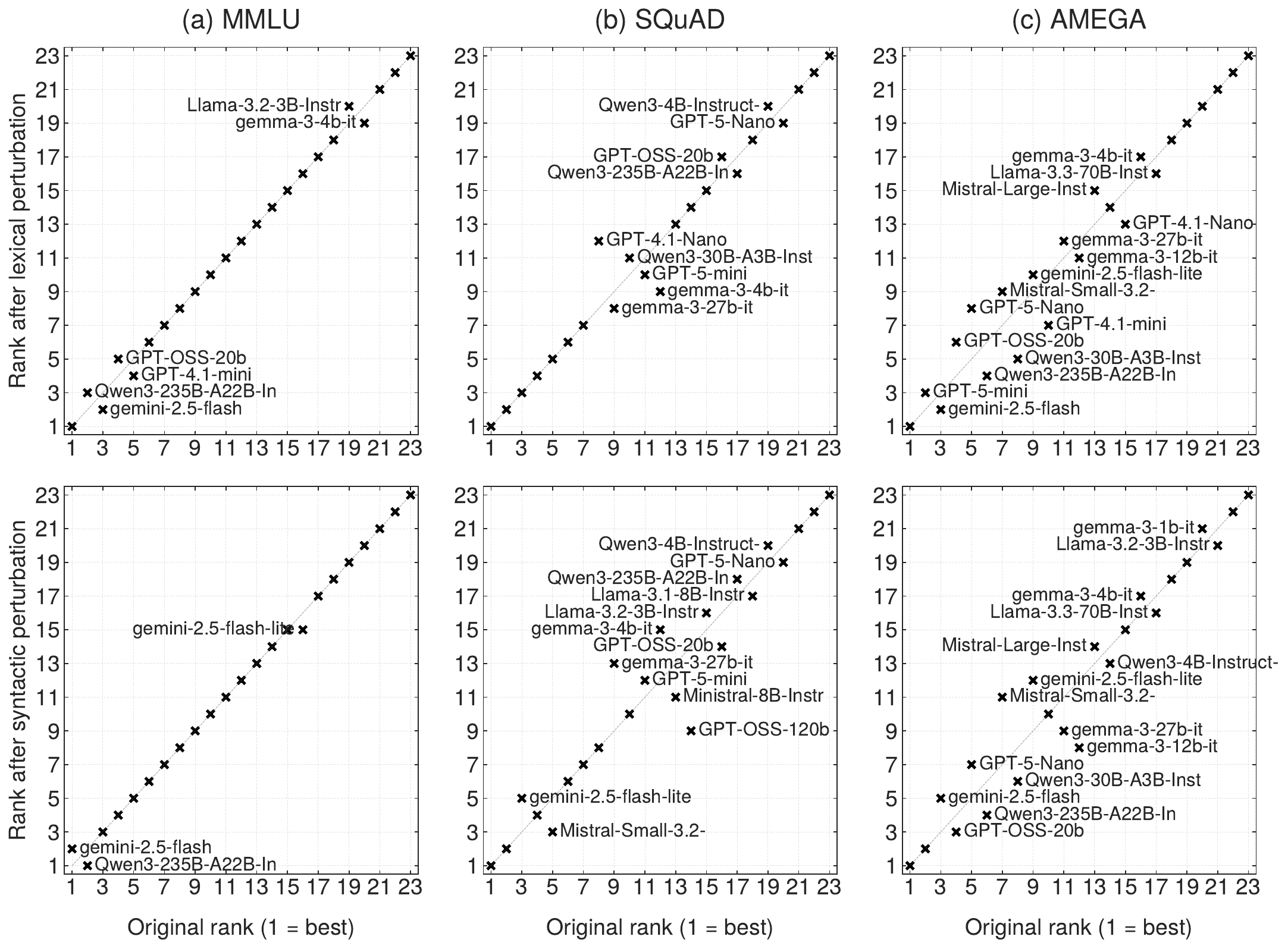}
    \caption{Model performance rankings before and after lexical perturbation (top) and syntactic perturbation (bottom) for (a) MMLU, (b) SQuAD, and (c) AMEGA. Rankings are largely preserved on MMLU, while SQuAD and AMEGA show noticeably more movement.}
    \label{fig:ranks}
\end{figure*}

Comprehensive per-model results are presented in Table~\ref{tab:results}, listing for each model the original score on MMLU, SQuAD, and AMEGA, along with absolute changes under lexical and syntactic perturbations.
Our analysis reveals three key findings regarding LLM robustness to meaning-preserving linguistic variations.

\newpage
\paragraph{Finding 1: LLMs are More Sensitive to Lexical Variation.}
A comparison of model performance on original and perturbed versions reveals that lexical perturbations consistently result in a statistically significant degradation of performance across nearly all models and benchmarks. In contrast, syntactic transformations have a much subtler impact.
As shown in Figure~\ref{fig:avg_performance_drop}, the performance drop induced by lexical changes is substantially larger than that from syntactic changes across all three datasets.

On MMLU, lexical perturbations caused an average accuracy drop of 7.72 percentage points (pp), compared to just 1.64 pp for syntactic changes.
This pattern holds for SQuAD (3.38 pp drop in F1 for lexical perturbation vs. 2.72 pp for syntactic perturbation) and AMEGA (1.25 point drop in adherence score for lexical perturbation vs. 0.54 for syntactic perturbation), although it is less pronounced.
An intra-model analysis about systematic differences using McNemar's test confirms that the performance drops from lexical perturbation are highly significant ($p<0.001$ for 22 of 23 models on MMLU), while the smaller drops from syntactic perturbation are also often significant but of a lesser magnitude. 
This discrepancy suggests that models' sensitivity is more strongly tied to surface lexical patterns than to abstract grammatical structures.

\paragraph{Finding 2: Leaderboards are Unstable.}
\begin{table}
\centering
\small
\setlength{\tabcolsep}{3pt}
\renewcommand{\arraystretch}{1.05}
\begin{tabular}{llccc}
\toprule
\textbf{Perturb.} & \textbf{Statistic} & \textbf{MMLU} & \textbf{SQuAD} & \textbf{AMEGA} \\
\midrule
\multirow{5}{*}{Lexical}
& Kendall's $\tau$                 & 0.98               & 0.93               & 0.89 \\
& 95\% CI (lower)                    & 0.93               & 0.83               & 0.80 \\
& \makecell[l]{Strict agreement\\($\tau>0.9$)}       & \Checkmark$^{*}$   & \XSolid            & \XSolid \\
& \makecell[l]{Mod.~agreement\\($\tau>0.8$)}      & \Checkmark$^{*}$   & \Checkmark$^{*}$   & \Checkmark$^{*}$ \\
& Score shift              & \Checkmark$^{***}$ & \Checkmark$^{***}$ & \Checkmark$^{***}$ \\
\midrule
\multirow{5}{*}{Syntactic}
& Kendall's $\tau$                  & 0.99               & 0.87               & 0.87 \\
& 95\% CI (lower)                   & 0.97               & 0.75               & 0.75 \\
& \makecell[l]{Strict agreement\\($\tau>0.9$)}         & \Checkmark$^{*}$   & \XSolid            & \XSolid \\
& \makecell[l]{Mod.~agreement\\($\tau>0.8$)}        & \Checkmark$^{*}$   & \XSolid            & \XSolid \\
& Score shift              & \Checkmark$^{***}$ & \Checkmark$^{***}$ & \Checkmark$^{**}$  \\
\bottomrule
\end{tabular}
\caption{Comparison of rankings on original vs.~perturbed MMLU, SQuAD, and AMEGA datasets using Kendall's $\tau$ and Wilcoxon signed-rank test. Asterisks denote significance ($^{*}\!:\!p<0.05$, $^{**}\!:\!p<0.01$, $^{***}\!:\!p<0.001$). Rankings are stable on MMLU but weaken on SQuAD and AMEGA.}
\label{tab:rank_corr}
\end{table}

Beyond absolute performance, perturbations alter the relative ranking of models, challenging the reliability of leaderboards. 
Figure~\ref{fig:ranks} illustrates the rank shifts for both perturbation types. 
For MMLU, model ranks remain highly stable under both lexical (a) and syntactic (d) perturbations, with most points clustered around the diagonal. 
In contrast, the plots for SQuAD (b, e) and AMEGA (c, f) exhibit more pronounced rank volatility.
For example, under lexical perturbation on SQuAD, GPT-4.1-Nano falls from 8th to 12th rank.
Similarly, under syntactic perturbation on SQuAD, GPT-OSS-120b improves its rank from 14th to 9th.

This observation is quantified using Kendall's rank correlation coefficient ($\tau$), as detailed in Table~\ref{tab:rank_corr}. 
For lexical perturbations, the rank correlation for MMLU is very high ($\tau=0.98$), meeting the threshold for strict equivalency and indicating high stability. 
Conversely, the correlations for SQuAD ($\tau=0.93$) and AMEGA ($\tau=0.89$) are demonstrably weaker. This reduced stability is confirmed by bootstrapped confidence intervals, which show that these models meet only the moderate agreement threshold for lexical perturbations.
A similar pattern emerges for syntactic perturbations: MMLU rankings are exceptionally stable ($\tau=0.99$), while SQuAD ($\tau=0.87$) and AMEGA ($\tau=0.87$) rankings are significantly less so, failing to meet the threshold for moderate equivalency.
Complementing the rank analysis, a Wilcoxon signed-rank test on per-model scores shows that the score distributions differ significantly between original and perturbed sets for both lexical and syntactic variation on all assessed benchmark datasets, confirming a systematic distributional shift.
These shifts suggest that leaderboards for complex tasks are brittle and sensitive to linguistic variation.

\begin{figure*}
    \centering
    \includegraphics[width=\linewidth]{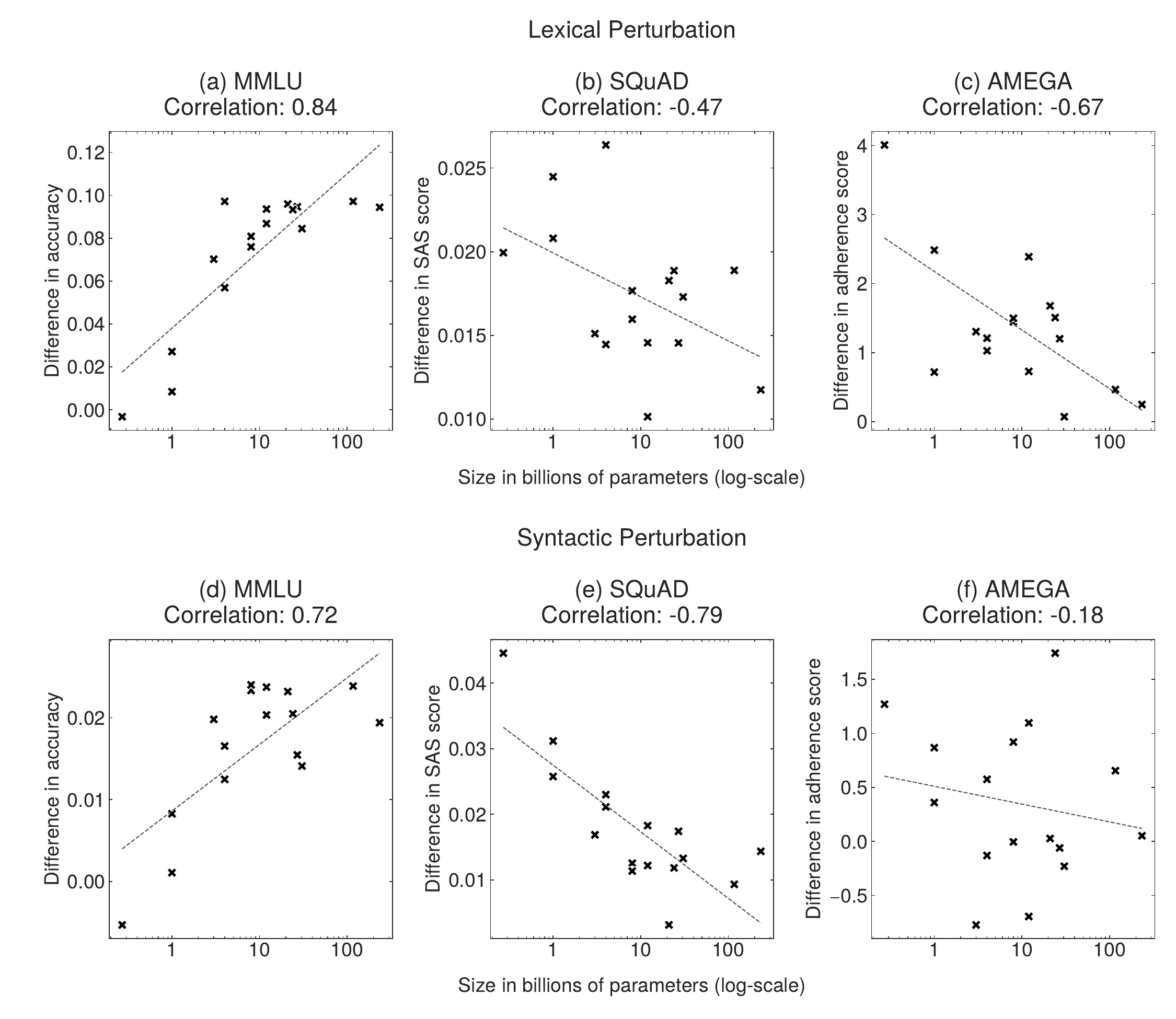}
    \caption{Correlation between log-transformed model size and performance drop on lexically and syntactically perturbed benchmarks (MMLU, SQuAD, and AMEGA). The dashed line in each plot illustrates the Ordinary Least Squares regression fit for the points. Model size correlates positively with performance drop on MMLU and negatively on SQuAD and AMEGA.}
    \label{fig:size_correlation}
\end{figure*}

\paragraph{Finding 3: Bigger $\neq$ More Robust.}
The common assumption that larger models are universally more robust is directly contradicted by our data.
The relationship between a model's parameter count and its robustness to perturbation is highly task-dependent.
Figure~\ref{fig:size_correlation} plots the performance drop against model size (log-scale) for lexical and syntactic perturbations.

For the MMLU benchmark, there is a strong positive correlation ($r=0.84$) between model size and the magnitude of the accuracy drop, suggesting that larger models are less robust on this task.
Conversely, for SQuAD ($r=-0.47$) and AMEGA ($r=-0.67$), the correlation is negative, indicating that for these tasks, larger models tend to be more robust.
A similar task-dependent pattern was observed for syntactic perturbations, with a positive correlation on MMLU ($r=0.72$) and a strong negative correlation on SQuAD ($r=-0.79$).
This complex relationship demonstrates that scale does not universally confer stability.
Its effect is rather contingent on the nature of the task.

\section{Conclusion}
In this paper, we systematically investigated 23 LLMs against linguistically principled, meaning-preserving perturbations across three diverse benchmarks.
Our findings reveal a clear hierarchy of sensitivity: LLMs are significantly more vulnerable to lexical substitutions than to syntactic transformations.
This result hints that their performance is more reliant on surface-level word patterns than on an abstract understanding of grammatical structure.

Furthermore, we demonstrated that both perturbation types are sufficient to destabilize benchmark leaderboards, especially on complex tasks, such as extractive question-answering and guideline-based reasoning.
This finding directly challenges the reliability of static leaderboards as the sole instrument for model selection.
Our results suggest that a top-scoring model might just be the best at overfitting to the benchmark's specific language.

Moreover, our analysis refutes the common assumption that model scale universally confers robustness.
Our findings show that "bigger is better" is a fallacy regarding stability. 
The effect of size is task-dependent, and in some cases, greater scale can even be detrimental, with larger models showing greater fragility on MMLU but increased robustness on SQuAD and AMEGA.

\paragraph{Design Challenges of Model Leaderboards.} Taken together, our results underscore the need to move beyond static performance metrics toward a more veridical evaluation paradigm.
The findings demonstrate that model leaderboards, particularly for complex tasks such as AMEGA, are fragile and can be destabilized by both lexical and syntactic variations.
This instability challenges the common practice of selecting models based on slight differences in reported benchmark scores, indicating that robustness evaluation should become a standard component of LLM assessment.
For practitioners, the observed sensitivity to lexical choice is a critical vulnerability in high-stakes applications. 
Therefore, we advocate for the integration of robustness testing into standard evaluation protocols to yield a more reliable assessment of LLM capabilities.
To support this effort, we release our perturbed benchmark datasets and implementation, encouraging further research into the linguistic generalization of language models.

\paragraph{Future Work.}
Our findings surface several questions for future research.
A primary direction is to investigate the mechanisms behind the observed effects, including the greater sensitivity of LLMs to lexical versus syntactic changes and the task-dependent link between model size and robustness.
Probing these mechanisms can clarify whether models rely on memorized surface patterns or develop more abstract reasoning.
Additionally, future research should broaden the scope of this investigation by diversifying the linguistic perturbations and applying them across a wider range of models, tasks, languages, and modalities.
Moreover, the research agenda should shift from merely diagnosing model fragility to proactively engineering for robustness.
This process involves leveraging interpretability techniques to gain a deeper understanding of failure modes, which can then guide the creation of novel training methodologies.
For example, strategies such as data augmentation with perturbed inputs or the use of a semantic consistency loss, which penalizes models for differing outputs on synonymous inputs, might directly lead to more stable and reliable systems.

%\iffalse
\section{Acknowledgements}
Bogdan Kostić and Julian Risch are supported by the European Union’s Horizon Europe Framework under Grant Agreement No.\ 101213369 \euflag\ as part of the DVPS project.

Additionally, this work is funded by the German Federal Ministry of Education and Research (BMBF) under the grant agreements 01|S23015A (\textbf{AI4SCM}) and with the project \textbf{SOOFI}: Large Reasoning Models, Grant-ID 13IPC040D, by Federal Ministry of Economic Affairs and Climate Action. 
This work is also funded by the Deutsche Forschungsgemeinschaft (DFG, German Research Foundation) Project-ID 528483508 -- \textbf{FIP 12}, as well as the European Union under the grant project 101079894 (\textbf{COMFORT} -- Improving Urologic Cancer Care with Artificial Intelligence Solutions). 
Views and opinions expressed are however those of the author(s) only and do not necessarily reflect those of the European Union or European Health and Digital Executive Agency (HADEA). 
Neither the European Union nor the granting authority can be held responsible for them.
%\fi

\section{Bibliographical References}\label{sec:reference}

\bibliographystyle{lrec2026-natbib}
\bibliography{bib}

\clearpage
\onecolumn
\appendix
\section{Syntactic Transformations and Applicability Conditions}
\label{sec:appendix_syntactic_transformations}
\begin{table}[htpb]
\centering
\begin{tabular}{l p{11cm}}
\toprule
\textbf{Transformation} & \textbf{Required Conditions} \\
\midrule
\textbf{Active to} & 
    \tabitem A nominal subject (\texttt{nsubj}).\\
    \textbf{Passive} &\tabitem A direct object (\texttt{dobj}).\\
    &\tabitem The subject is not the pronoun `\emph{it}'.\\
    &\tabitem The main verb is not `\emph{have}'. \\
\addlinespace
\hline
\addlinespace
\textbf{Passive to} & 
    \tabitem A passive nominal subject (\texttt{nsubjpass}).\\
    \textbf{Active} &\tabitem A passive auxiliary verb (\texttt{auxpass}).\\
    &\tabitem An agent (\texttt{agent}) phrase.\\
\addlinespace
\hline
\addlinespace
\textbf{Extraposition} & 
    \tabitem A clausal subject (\texttt{csubj}).\\
\addlinespace
\hline
\addlinespace
\textbf{Reverse } & 
    \tabitem The nominal subject (\texttt{nsubj}) is the pronoun `\emph{it}'.\\
   \textbf{Extraposition} &\tabitem A clausal complement (\texttt{ccomp}).\\
\addlinespace
\hline
\addlinespace
\textbf{Wh-Movement} & 
    \tabitem A Wh-word that is not the subject.\\
    &\tabitem The subject appears before the auxiliary verb (if any).\\
\addlinespace
\hline
\addlinespace
\textbf{Reverse} & 
    \tabitem A Wh-word that is not the subject.\\
    \textbf{Wh-Movemen}t &\tabitem An auxiliary verb appears before the subject.\\
\addlinespace
\hline
\addlinespace
\textbf{Dative Alternation} & 
    \tabitem A direct object (\texttt{dobj}).\\
    &\tabitem A nominal dative indirect object (\texttt{dative}).\\
\addlinespace
\hline
\addlinespace
\textbf{Prep.~Dative} & 
    \tabitem A direct object (\texttt{dobj}).\\
   \textbf{Alternation} &\tabitem A prepositional dative indirect object (\texttt{dative}), e.g., `\emph{to him}''.\\
\bottomrule
\end{tabular}
\caption{Complete list of syntactic transformations and their corresponding rule-based applicability conditions used in the syntactic perturbation pipeline. Conditions rely on dependency parsing tags.}
\label{tab:applicability_conditions}
\end{table}

\end{document}